# Solving The Travelling Salesmen Problem using HNN and HNN-SA algorithmn

Gyanateet Dutta

School of Electronics Engineering

Kalinga Institute of Industrial Technology

Bhubaneshwar , Odisha , India

1930198@kiit.ac.in

*Abstract-*
**In this case study, the renowned Travelling Salesmen problem has been studied. Travelling Salesmen problem is a most demanding computational problem in Computer Science. The Travelling Salesmen problem has been solved by two different ways using Hopfield Network. The main theory of the problem is to find distance and connectedness between nodes in a graph having edges between the nodes. The basic algorithm used for this problem is Djikstra's Algorithm. But till now, a number of such algorithms has evolved. Among them (some other algorithms) , are distinct and have been proved to solve the travelling salesmen problem by graph theory.**

*Keywords: Neural network , Travelling Salesmen Problem, Hopfield Network, Simulated Annealing.*

## 1. Introduction

The travelling salesman problem (TSP) is an algorithmic problem tasked with finding the shortest route between a set of points and locations that must be visited. In the problem statement, the points are the cities a salesperson might visit. The salesman's goal is to keep both the travel costs and the distance travelled as low as possible.

TSP has been studied for decades and several solutions have been theorized. The simplest solution is to try all possibilities, but this is also the most time consuming and expensive method. Many solutions use heuristics, which provides probability outcomes. However, the results are approximate and not always optimal. Other solutions include branch and bound, Monte Carlo and Las Vegas algorithms.

Following are the constraints of a TSP problem :
1. The total length of the loop should be a minimum.
2. The salesperson cannot be at two different places at particular time.

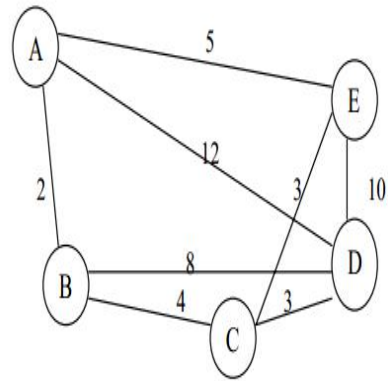

Figure 1 : A graph with weights on its edges.

For example, both Path1 (ABCDEA) and Path2 (ABCEDA) passes through all vertices . However it can be seen that the total length of Path1 is 24, where as the total length of Path2 is 31

### 1.1 History of TSP

A mathematician named Karl Menger, first studied T he TSP problem in its general form in Vienna in the year 1920. In 1954, G. Dantzig et al. solved a non-trivial TSP with 49 cities. As the years passed, the number of cites for which the TSP has been solved gradually increased. Even though the work performed on TSP is often not directly applicable, it provides a platform to investigate general methods to solve discrete optimization problems.Even though the work performed on TSP is often not directly applicable, it provides a platform to investigate general methods to solve discrete optimization problems. TSP poses a problem in many transportation and logistics based applications. For example, an application to arrange school routes such that all the children are picked up from a particular school district. This application has proved to be a historical significance to TSP, as it provided a motivation for Merrill Flood during the 1940s. In many industries, there has been a great concern regarding the methods to be applied to increase the efficiency of the processes. This is because many of these processes involve, time consuming tasks of moving items from one location to another. It can be observed that many commercial based applications have been developed in this field using the concept of TSP.

## 1.2 Simulated Annealing

By definition, Simulated Annealing is a generic probabilistic meta-algorithm used to find an approximate solution to global optimization problems. Simulated Annealing is inspired by the concept of annealing in metallurgy which is a technique of controlled material cooling used to reduce defects (M. Sureja and V. Chawda, 2012) .

To use Simulated Annealing, the system must first be initialised with a particular configuration. The SA algorithm then starts with a random solution. With every iteration, a new random nearby solution is formed. If the solution is better, then it will replace the current solution. If it is worse, then based on the probability of the temperature parameter, it may be chosen to replace the current solution. As the algorithm progresses, the temperature tends to decrease, thus making sure that worse solutions have a lesser chance of
replacing the current solution.

*Pseudo Code (Wikipedia) :*

```
Let s = s0
For k = 0 through kmax (exclusive):
T ← temperature(k
∕kmax)
Pick a random neighbour, snew ←
neighbour(s)
If P(E(s), E(snew), T) ≥ random(0, 1),
move to
the new state:
s ← snew
Output: the final state s
```

As per Figure 2, if the energy of the new state is less than that of its previous one, then the change is accepted unconditionally and the system is updated. But if the energy is greater, than the latest configuration has the probabilistic chance or being accepted based on the temperature parameter.

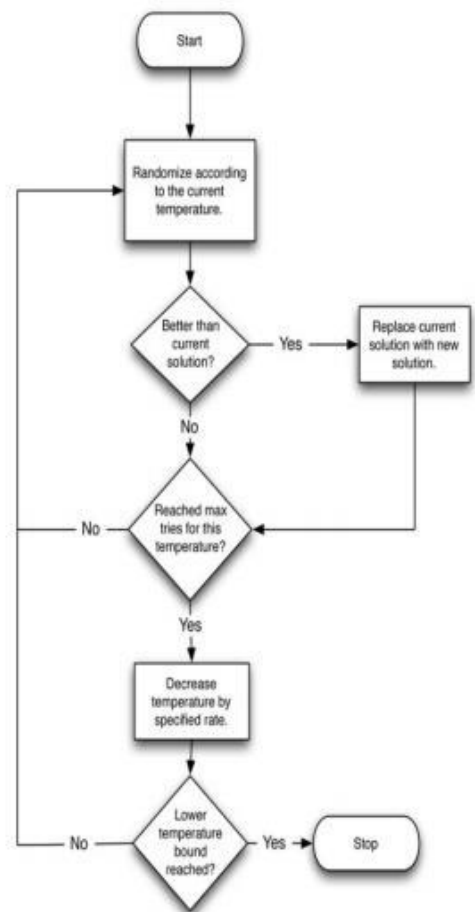

Figure 4 : Flowchart of Simulated Annealing

But if the energy is greater, than the latest configuration has the probabilistic chance or being accepted based on the temperature parameter.

The remainder of the paper is explained as follows,

Section 2 – Explains about the aim of the paper and a list of objectives that it will be covering.
Section 3 – Provides a basic comparison of various TSP Algorithms.
Section 4 – Contains details of the case study on TSP algorithm using Hopfield neural network and
Simulated Annealing. It provides an insight to both theoretical and practical implementations of the
algorithm using MATLAB.
Section 5 – Discusses about the various articles published over time.
Section 6 – Contains a list of references used in this paper.

# 2.Related Work

Jacek Mańdziuk[1] proposed a modification of the Hopfield neural network for solving the Travelling Salesmen Problem by using the calculating the output potential as
$v(u) = 1/2 \ [1 + \tanh(\alpha u)]$ where $\alpha$ is the gain of the function.

Sarat Kumar Sarvepelli[2] proposed a HNN and SA network for solving the Travelling Salesmen Problem.

Prof, Sharadindu Roy, Prof Samer Sen Sarma, Soumyadip Chakravorty , Suvodip Maity proposed an HNN method for solving TSP

# 3.Estimation Techniques

There were many heuristic based algorithms developed across a period of time to obtain near-optimal solutions. Some of these algorithms are as follows:
1)Greedy Algorithm.
2)Simulated Annealing.
3)2-opt Algorithm
4)3-opt Algorithm.
5)Genetic Algorithm.
6)Neural Network(HNN) etc.

## 3.1 Greedy Algorithm
 Greedy algorithm is the simplest improvement algorithm by definition.
The algorithm starts with a Node numbered 1 and calculates all the distances to other (n-1) nodes . Then, it goes to the closest node.Considering the current node as the departure node , it then selects the next nearest node from the (n-2) remaining nodes.This process continues until every node is visited atleast once and then goes back to Node 1.(Kim,Shim and Zhang, 1998)

## 3.2 Simulated Annealing
Simulated Annealing(SA), by definition, is a generic probabilistic meta-algorithm designed to find an appropriate solution to global problems based on optimization.
Simulated Annealing is inspired by the concept of annealing in metallurgy , which is a technique of controller material cooling used to reduce defects(M.Sureja and V.Chawda) 2012.

To be able to use the SA  , the system must be first intialized with a particular configuration. Then the SA algorithm starts with a random solution. With every iteration, a new random solution is formed. If the solution turns out to be better, it replaces the current solution. If it turns out worse, then based on the probability of the temperature parameter, it may be chosen to replace the current solution. As the algorithm progresses, the temperature parameter tends to decrease, thus making sure that worse solutions have a lesser chance of replacing the current solution. As the algorithm progresses, the temperature tends to decrease, thus making sure the worse solutions have a lesser chance of replacing the current solution.

## 3.3 2-opt algorithm
Croes first propsed the 2-opt in 1958 as a simple local search algorithm for solving the travelling salesman problem. The main idea behind the 2-opt algorithm was to take a route which crosses over itself and reorder it so that it does not. The results of the 2-Opt algorithm was first published in 1956.

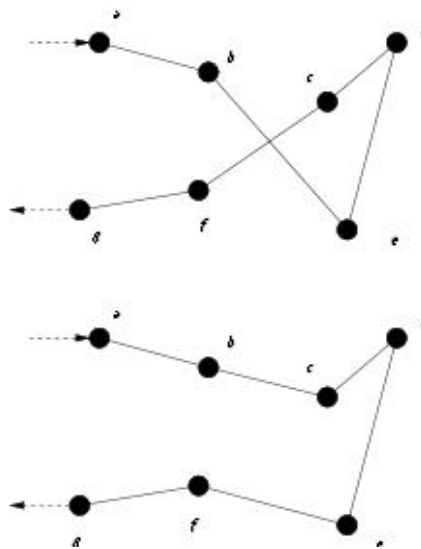

Figure- 2-Opt Algorithm

## 3.4 3-opt algorithm

3-opt algorithm involves removing 3 connections (or edges) in a network(or tour) and then reconnecting the network in all other possible ways by evaluating each reconnection to find the optimum one. The process is later repeated for a different set of 3 connections.

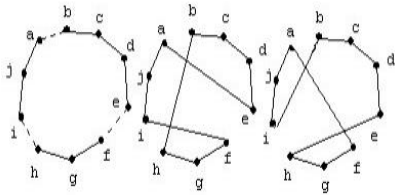

Figure 6- 3-opt Algorithm

Except for the difference in splitting , the 3-Opt algorithm is quite similar to the 2-Opt algorithm.These two are in fact part of a family of methods called K-Op.

## 3.5 Genetic Algorithm

By definition, Genetic Algorithm is an optimization method that is based on the biological analogy of "survival of the fittest". By the process of genetic mutation , crossover and genetic reproduction , the quality of an average population and its individuals is improved across several generations.

## 3.6 Hopfield Neural Network

In 1983, a physicist John Hopfield published the famous paper "Neural network and physical system with emergent collective computational abilities". Along with the rediscovery of Backpropagation and the introduction of cheap computing power, this helped to reignite the dormant world of Neural Networks once again.

By definition, A Hopfield artificial neural network is a type of recurrural network artificial network that is used to store one or more stable vectors. These stable vectors can be viewed as memories that the network recalls when provided with similar vectors that can act as a cue to the network memory. The binary units only take two different values for their states that are determined by whether or not the unit input exceeds their threshold. Binary unit can take either values of 1 or -1, or values of 1 or 0. Consequently there are two possible definitions for binary unit activation.

### 3.6.1 Basic Operation

Hopfield was to add feedback connection to the network( the outputs are fed back into the inputs) and show that with these connections the network are capable of interesting behaviors which we might not expected of them, in particular they can holds memories. Networks with such connections are called "feedback" or "recurrent" networks.

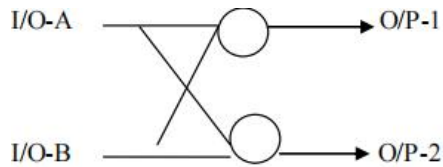

*Fig: Free Forward Network*

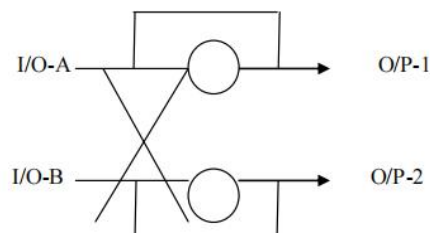

*Fig: Free Forward Network with Feedback* connection

The neural network for any application can be best understood by its energy function.

The energy function contains various sections that represent the patterns stored in the network. The input pattern of the network, represents a particular point in the energy landscape. As the input pattern iterates its way to a solution, the point slowly moves towards one of the sections of the landscape. The iterations are either carried out for a fixed number of time or until a stable state is reached. (Graupe,2007)

The energy function in a HNN network should satisfy the following conditions:

- The function should lead to a stable state.
- It should provide the shortest travelling path.

The neurons in the network must also satisfy the following criteria (Hopfield networks - Comp Dept Leeds) :
1. The value of each input, xi is determined and the weighted sum of all inputs is $\sum_i w_i x_i$ calculated.

2. The output state of the neuron is set to +1 if the weighted input sum is larger or equal to 0. It

is set to -1 if the weighted input sum is smaller than 0.

3. A neuron retains its output state until it is updated again.

$$o = \begin{cases} 1 & : \sum_i w_i x_i \geq 0 \\ -1 & : \sum_i w_i x_i < 0 \end{cases}$$

The remainder of the paper is explained as follows,

Section 4 – Contains details of the case study on TSP algorithm using Hopfield neural network and SA.

Section 5 – Contains results of the case study.

Section 6 – Contains conclusion of the case study

Section 5 – Contains a list of references used in this paper.

Rather than focus on finding the most effective route, TSP is often concerned with finding the cheapest solution. In TSPs, the large amount of variables creates a challenge when finding the shortest route, which makes approximate, fast and cheap solutions all the more attractive.

Monte Carlo Simulation, also known as the Monte Carlo Method or a multiple probability simulation, is a mathematical technique, which is used to estimate the possible outcomes of an uncertain event. The Monte Carlo Method was invented by John von Neumann and Stanislaw Ulam during World War II to improve decision making under uncertain conditions. It was named after a well-known casino town, called Monaco, since the element of chance is core to the modeling approach, similar to a game of roulette.

Las Vegas algorithms are nondeterministic algorithms with the following properties: If a solution is returned, it is guaranteed to be correct, and the run-time is characterized by a random variable. Las Vegas algorithms are prominent not only in the field of Artificial Intelligence but also in other areas of computer science and Operations Research.

# 4. My contribution

In this paper I have analysed two solution methodologies in detail and the way they both work.

**4.1 Solution Methodologies**

**4.1.1 TSP using Hopfield Neural Network(HNN)**

Using the continuous Hopfield Neural Network it is possible to find a solution to this problem. Here in this figure the fully connected weighted network with the output of the network fed back and its weights to each link are shown.

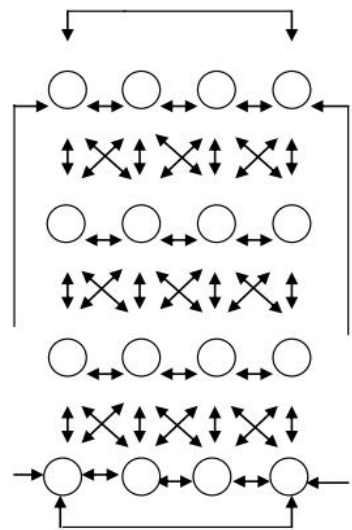

$N^2$ neurons are used in the network, where n is the total number of cities in the problem. The neurons here have a threshold and step function. The inputs are given to the weighted nodes. The network then would calculate the output and based on the energy function and weight update function would converge to the stable solution after few iteration. However the essential task is to find an appropriate connection weight. That should be such that invalid tour would be prevented and valid tour would be preferred.

The example here has been taken for 4 cities. The 4 cities TSP would need 16 neurons. The output result of TSP can be represented as:

|   | 1 | 2 | 3 | 4 |
|---|---|---|---|---|
| A | 0 | 1 | 0 | 0 |
| B | 1 | 0 | 0 | 0 |
| C | 0 | 0 | 0 | 1 |
| D | 0 | 0 | 1 | 0 |

Figure- Tour Matrix obtained as the output of the network.

The corresponding visiting route , in the example is
City B   City A   City D   City C   City B.
So the total distance would be D= AB+BC+DC+DA

### 4.1.1.1 Input To The Network

The inputs to the network are chosen arbitrarily. The initial state of the network is not fixed and is biased against any particular route. As a consequence of the choice of the inputs, the activation function gives out outputs that add up to the number of cities and the initial solution for the problem , a legal tour will be the result. Also there are inputs that would be taken from the user. The user would be asked to input the number of cities which would be used to generate the distance matrix .

The distance matrix is a n*n square matrix whose principal diagonal is zero

|   | 1  | 2  | 3  | 4  |
|---|----|----|----|----|
| A | 0  | 15 | 13 | 17 |
| B | 15 | 0  | 14 | 27 |
| C | 13 | 14 | 0  | 25 |
| D | 17 | 27 | 25 | 0  |

Fig: Distance matrix generated after getting inputs from the user.

In the example , the distance between city A and city C is 13 and distance between a city itself is zero.

### 4.1.1.2 Energy Function

The Hopfield network for the applying a neural network is best  understood by the energy function. The energy function developed by Hopfield and Tank is used. The energy function has various hollows that represent the patterns stored in the network. An unknown input pattern represents a particular point in the energy landscape and the pattern iterates its way to a solution, the point moves through the landscape towards one of the hollows.This iteration is carried on till some fixed number of times or till a stable state is reached.

The energy function should satisfy the following criterion:

- The energy function should be able to lead to a stable  combination matrix
- The energy function should lead to the shortest travelling path.

The energy function used for this example of HNN is

$E = A/2 \sum_{nx=1} \sum_{ni=1} \sum_{nj=1, j \neq I} v_{xi}v_{xj}$
$+ B/2 \sum_{ni=1} \sum_{nx=1} \sum_{ny \neq x} v_{xi}v_{yi}$
$+ C/2 (\sum_{nx=1} \sum_{ni=1} v_{xi} - n)^2$
$+ D/2 \sum_{nx=1} \sum_{ny=1, y \neq x} \sum_{ni=1} d_{xy} v_{xi}(v_{y,i+1} + v_{y,i-1})$

where, A,B,C,D are the constants which are to be determined by trial and error method.

If all $v_{xi}$ approach either 0 or 1 and if those with $v_{xi} \approx 1$ represents a Hamiltonian circuit then $d_{xy}v_{xi}(v_{y,i+1}+v_{y,i-1}) = d_{xy}$ (if city y is either before or after city x in the circuit)  = 0 otherwise.

### 4.1.1.3 Row Term

$(A/2 \sum_{nx=1} \sum_{ni=1} \sum_{nj=1, j \neq I} v_{xi}v_{xj})$ in the energy function , the first triple sum is zero if and only if there is only one "1" in each order column. Thus this takes care that no two or more cities are in same travel order. i.e. no two cities are visited simultaneously.

### 4.1.1.4 Column Term

$(B/2 \sum_{ni=1} \sum_{nx=1} \sum_{ny \neq x} v_{xi}v_{yi})$ in the energy function , the 2nd  triple sum is zero if and only if there appears only one city  in each order column. Thus this takes care that each city is visited only once.

### 4.1.2.5 Total number of "1" term

$(C/2 (\sum_{nx=1} \sum_{ni=1} v_{xi} - n)^2)$ in the energy function , the 3rd triple sum is zero if and only N numbers of 1 appear in the whole n*n matrix . Thus this takes into care that all the cities are visited.
The first three summations are to satisfy condition 1, which is necessary to produce a legal travelling path.

### 4.1.1.6 Shortest distance term

$(D/2 \sum_{nx=1} \sum_{ny=1, y \neq x} \sum_{ni=1} d_{xy} v_{xi}(v_{y,i+1}+v_{y,i-1}))$ the 4th summation provides the term for the shortest path.
D is the distance between city i and city j , the value of this term is minimum when the total distance travelled is shortest

### 4.1.2 THE HOPFIELD TSP ALGORITHM

Step 1: Initializing n , i.e the number of cities ; D[n x n] the distance matrix ; I[n x n] the identity matrix and the tour variable is initialized as 0.

Step 2: Using identity matrix i number of tour matrices are created where tour matrix is T[n x n]. And the total four matrices are S=n! . P[s]= T[n x n]

Step 3: for j=1 upto s

Step 4: P[s]

Step 5: for x=1 upto n

Step 6: for y=1 upto n
         Set i=y

Step 7: if T[x][1] equal to 1 then
         Set s1=x;

Step 8: if T[x][n] equal to 1 then
         Set s1=x;

Step 9: if i-1 equal to 0 then
         Set i-1 =s1;

Step 10: if i+1 garter then n then
         Set i+1 = s2

Step 11: d=D[x][y] × T[x][i] (T[y][i+1]+T[y][i-1])

Step 12: tour= tour+d

Step 13: end

Step 14: end

Step 15: tours[k]=tour

Step 16: end

Step 17: set TSP=find(tour==min(tour))

Step 18: end

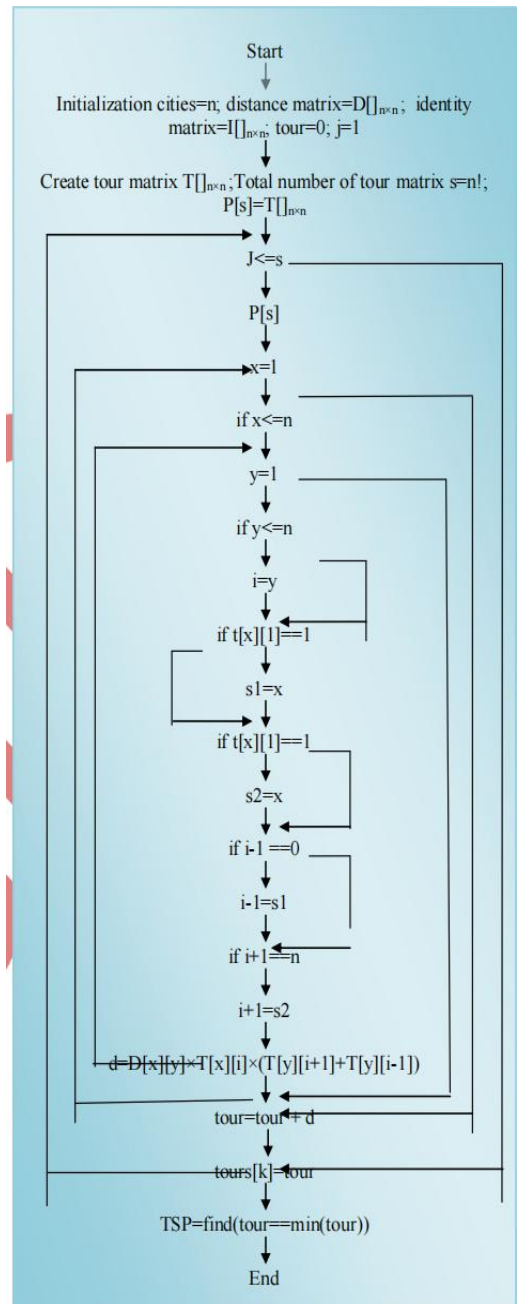

### 4.1.3 TSP Using Hopfield Neural Network and Simulated Annealing.

#### 4.1.3.1 Algorithm:

The algorithm used in this scenario follows a fully connected topology. The execution of the algorithm as follows:

Step 1: The number of cities and their respective position, distance between each other and corresponding weights are initialised.

Step 2: During the initialisation phase the location of the cities are provided as an input to the Simulated Annealing Algorithm.

Step3: The output locations of the SA algorithm are then fed into the Hopfield NN Algorithm as the input locations

Step4: The HNN algorithm is then executed to make sure to get the optimised path for the salesman to travel.

4.1.3.2 Theoritical Solution:

**Scripts**
initTsp
ForwardHopfield

**Functions:**
Simulatedannealing()
Distance()
Plotcities()
Swapcities()
setWeight()

### 4.1.3.2.1 initTsp Script

This script is used to randomly create a number of cities in XY coordinate system. It is executed only once during which the variables required by the HNN algorithm are initialised .

For each city in the coordinate system , a respective stop is generated, I.e for :
C1,C2,C3,C4
S1,S2,S3,S4

Where C is the city and S is the stop. The total number of inputs (I) for HNN is the combination of the number of cities and stops.

The random locations of each city is then provided as an input to the simulated annealing algorithm I.e the simulatedannealing() function , along with its remaining parameters. The output of the algorithm is the initial path that a salesman can follow.

When the SA algorithm is completed once , the distance between each is calculated based on the coordinates. The IT is then normalised is such a way that the maximum distance between the cities would be equal to 1.0. Once the distance has been normalised , the weights of all inputs (I) is calculated which is stored in a weighted matrix using the setWeight() function.

### 4.1.3.2.2 Simulatedannealing()

This function has five input arguments(SESHADRI, 2006)

1. First argument - Array of cities

2. Second argument- Initial temperature
3. Third argument- Cooling rate
4. Fourth argument- Threshold , in this case the threshold is the total number of iterations. It can also be sent in terms of iterations. It can also be sent in terms of optimal solution or in terms of final temperature. However for the algorithm mentioned, it is set in terms of iterations.
5. Fifth argument- Number of cities to swap. Each time the cities are swapped, a random new configuration would be generated. From the current configuration , in order to get a new random configuration , either a pair of cities would be exchanged at random or else any number of pair of cities may be exchanged.

Initially , the total distance between the cities of an input order(I), is calculated using the distance() function and is stored in A.

### 4.1.3.2.3 Distance()

This function calculates the distance between n cities as it is required in a Travelling Salesmen Problem. The input argument has 2 rows and n columns , where n is the total number of cities and each column represents the location of the corresponding city (SESHADRI,2006)

### 4.1.3.2.4 Swapcities()

This function has two input parameters

- Input cities(I)
- Number of swaps(N)

It returns a set of "M" cities as output , where N cities are randomly swapped.

### 4.1.3.2.5 Plotcities()

The plotcities() function plots the function of cities based on the argument called inputcities. The argument has 2 rows and n columns, where n is the number of cities. Along with the location, the symmetric route between the cities is also plotted.

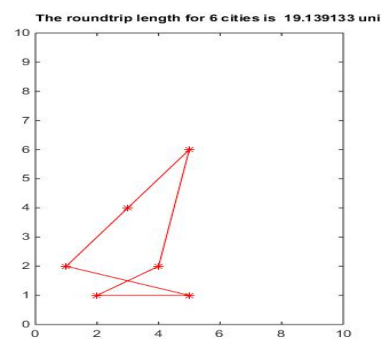

Figure : Output of SA algorithm

### 4.1.3.2.6 SetWeight()

The setweight() function changes the value of a single weight in the matrix to have the value 'wt'. Where 'wt' is "inhibWeight" for constraints I and II and negative distance for constraint III. The function also sets the value of symmetric weight.

The weight for each combination of city and stop , is stored in the weighted matrix as ,
C1S1    C1S2……
C2S1    C2S2……

### 4.1.3.2.7 ForwardHopfield Script

In this script, the input cities and stops are executed in a random order in the first cycle. For example, three cities can be executed in a random order( cities and stops) like:
6,1,4,3,5,2.

For each input unit in the order, a net weight(N) is calculated as follows,

Net Weight(N)= weight of the unit * activation matrix

It is then compared to the threshold value(V)

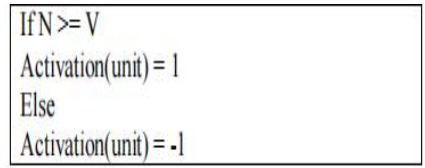

If the net weight is greater than the threshold value than the activation value for that unit is made 1.
Or else the value is made -1

## 5 Result section

### 5.1 HNN
Using the HNN algorithm , following results were obtained.

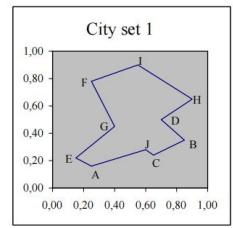

City Set 1 : cite coordinates and shortest tour

| PART | Best | Mean | Worst | % Succ. | Iter. | FULL | Best | Mean | Worst | % Succ. | Iter. |
|---|---|---|---|---|---|---|---|---|---|---|---|
| C=90, D=100 | | | | | | C=90, D=100 | | | | | |
| a | 2.785 | 3.19 | 3.69 | 100 | 80.7 | a | 2.767 | 3.21 | 3.79 | 100 | 121.1 |
| b | 2.696 | 3.22 | 3.65 | 100 | 64.4 | b | 2.765 | 3.21 | 3.88 | 100 | 107.0 |
| c | 2.696 | 3.17 | 3.68 | 100 | 64.4 | c | 2.696 | 3.21 | 3.70 | 100 | 97.1 |
| d | 2.696 | 3.20 | 3.63 | 100 | 51.1 | d | 2.696 | 3.21 | 3.69 | 100 | 106.3 |
| C=90, D=110 | | | | | | C=90, D=110 | | | | | |
| a | 2.696 | 3.00 | 3.31 | 100 | 129.8 | a | 2.765 | 3.01 | 3.34 | 97 | 219.8 |
| b | 2.696 | 2.99 | 3.31 | 100 | 127.1 | b | 2.696 | 2.99 | 3.41 | 95 | 238.2 |
| c | 2.696 | 3.00 | 3.33 | 100 | 165.1 | c | 2.696 | 3.02 | 3.43 | 96 | 223.4 |
| d | 2.696 | 3.01 | 3.33 | 99 | 154.7 | d | 2.696 | 3.04 | 3.52 | 96 | 231.6 |
| C=100, D=100 | | | | | | C=100, D=100 | | | | | |
| a | 2.877 | 3.52 | 4.38 | 100 | 33.4 | a | 2.937 | 3.49 | 4.08 | 97 | 51.8 |
| b | 2.765 | 3.51 | 4.21 | 100 | 34.5 | b | 2.787 | 3.53 | 4.51 | 95 | 65.5 |
| c | 2.785 | 3.54 | 4.42 | 100 | 31.4 | c | 2.785 | 3.45 | 4.14 | 93 | 68.7 |
| d | 2.696 | 3.55 | 4.20 | 100 | 33.9 | d | 2.767 | 3.47 | 4.15 | 97 | 61.4 |
| C=100, D=110 | | | | | | C=100, D=110 | | | | | |
| a | 2.696 | 3.24 | 3.88 | 92 | 107.3 | a | 2.696 | 3.22 | 3.89 | 82 | 157.3 |
| b | 2.696 | 3.26 | 3.89 | 93 | 113.5 | b | 2.765 | 3.27 | 4.05 | 79 | 159.2 |
| c | 2.696 | 3.26 | 3.88 | 93 | 137.4 | c | 2.696 | 3.22 | 3.79 | 78 | 155.4 |
| d | 2.696 | 3.23 | 3.88 | 94 | 109.9 | d | 2.696 | 3.25 | 3.75 | 81 | 116.3 |
| C=100, D=120 | | | | | | C=100, D=120 | | | | | |
| a | 2.785 | 3.06 | 3.42 | 54 | 223.5 | a | 2.696 | 3.06 | 3.46 | 42 | 224.6 |
| b | 2.696 | 3.04 | 3.46 | 59 | 162.7 | b | 2.696 | 3.05 | 3.46 | 34 | 231.2 |
| c | 2.696 | 3.01 | 3.46 | 52 | 201.3 | c | 2.696 | 3.05 | 3.41 | 36 | 208.0 |
| d | 2.696 | 3.03 | 3.46 | 66 | 210.0 | d | 2.696 | 3.10 | 3.58 | 44 | 219.2 |

Results of computer simulations for the city set 1 . Parameters: A=B=100, σ=1

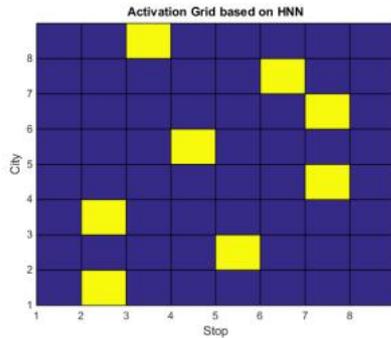

Figure : Activation Grid for 8 cities using only HNN

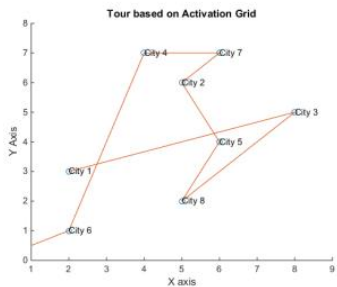

Figure: Tour for cities using only HNN

### 5.2 Using the HNN and SA algorithms together

Step1: Executing the initTsp Script

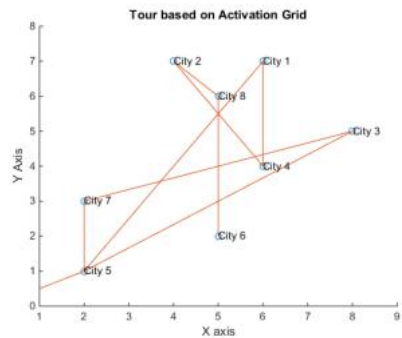

Now once the SA algorithm is executed , the optimised order of the cities is stored as an input to the HNN algorithm

Step2: Executing the forwardHopfield Script
This script is executed until the algorithm reaches a stable state where a message is displayed in the command window as "converged".

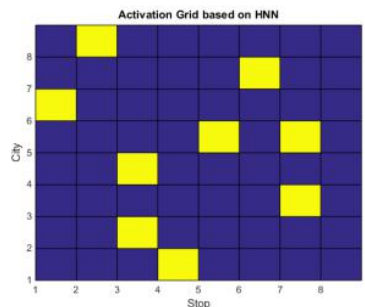

This would generate an activation grid and a plot for the optimised HNN tour

Figure: Activation Grid for 8 cities using HNN and SA

## 6 Conclusion
After finding a solution with each method i.e the HNN itself and the HNN combined with SA , it can be observed that the HNN algorithm can be sufficient enough to provide an optimum solution but as the number of cities increases , the HNN algorithm on its own fails to provide a good solution as the the input variables such as threshold and inhib weight need to be precisely specified to satisfy the constraints. In such cases , it is observed that by additionally including an SA algorithm during the initialisation helps to integrate local maps into a single radiation based hybrid map for a genome.